\documentclass[11pt]{article}

\usepackage[preprint]{acl}

\usepackage{times}
\usepackage{latexsym}
\usepackage{amsmath, amsfonts}
\usepackage{enumitem}
\usepackage{algorithm, algorithmic}
\usepackage{adjustbox}
\usepackage{multirow}

\usepackage[T1]{fontenc}
\usepackage[utf8]{inputenc}

\usepackage{microtype}

\usepackage{inconsolata}

\usepackage{graphicx}
\usepackage{tcolorbox}
\tcbuselibrary{breakable}
\usepackage{verbatim}
\usepackage{fvextra}
\usepackage{threeparttable}
\usepackage{mdframed}

\title{Asymmetric Actor–Critic for Multi-turn LLM Agents}

\author{
 \textbf{Shuli Jiang\textsuperscript{1}},
 \textbf{Zhaoyang Zhang\textsuperscript{1}},
 \textbf{Yi Zhang\textsuperscript{1}},
 \textbf{Shuo Yang\textsuperscript{1}},\\
 \textbf{Wei Xia\textsuperscript{1}},
 \textbf{Stefano Soatto\textsuperscript{1}}
\\
 \textsuperscript{1}AWS Agentic AI
\\
 \small{
    \{shulij, ozhaozha, yizhngn, shuoy, wxia, soattos\}@amazon.com
 }
}

\newcommand{\asymac}{Asym-AC}
\newcommand{\asymacsft}{Asym-AC-SFT}

\begin{document}
\maketitle
\begin{abstract}
% current challenge
    Large language models (LLMs) exhibit strong reasoning and conversational abilities, but ensuring reliable behavior in multi-turn interactions remains challenging. In many real-world applications, agents must succeed in \textit{one-shot} settings where retries are impossible. Existing approaches either rely on reflection or post-hoc evaluation, which require additional attempts, or assume fully trainable models that cannot leverage proprietary LLMs.
    We propose an \textit{asymmetric actor–critic} framework for reliable conversational agents. A powerful proprietary LLM acts as the actor, while a smaller open-source critic provides runtime supervision, monitoring the actor’s actions and intervening within the same interaction trajectory. Unlike training-based actor–critic methods, our framework supervises a fixed actor operating in open-ended conversational environments.
    The design leverages a generation–verification asymmetry: while high-quality generation requires large models, effective oversight can often be achieved by smaller ones. We further introduce a data generation pipeline that produces supervision signals for critic fine-tuning without modifying the actor.
    Experiments on $\tau$-bench and UserBench show that our approach significantly improves reliability and task success over strong single-agent baselines. Moreover, lightweight open-source critics rival or surpass larger proprietary models in the critic role, and critic fine-tuning yields additional gains over several state-of-the-art methods.
\end{abstract}

\section{Introduction}

\begin{figure}[t]
    \centering
    \includegraphics[width=\linewidth]{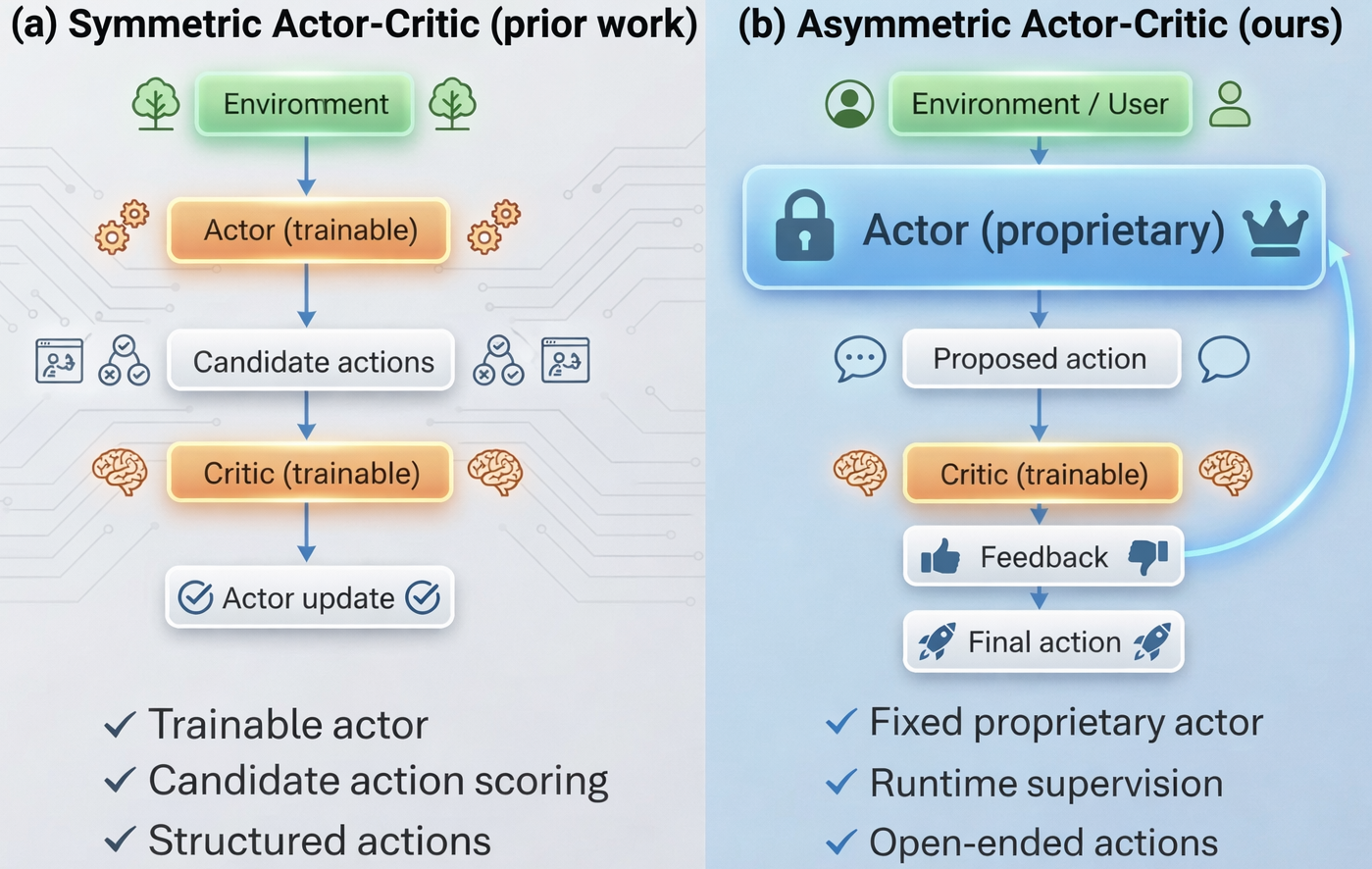}
    \caption{Existing actor–critic methods assume trainable actors and structured action spaces. We propose an asymmetric actor–critic framework where a lightweight open-source critic supervises a fixed proprietary actor at runtime, enabling reliable control for one-shot multi-turn conversational agents.}
    \vspace{-10pt}
    \label{fig:teaser}
\end{figure}

Large language models (LLMs) have demonstrated remarkable capabilities across a wide range of applications, including customer service~\cite{su2025llm_customer_support}, complex reasoning~\cite{xu2025largereasoningmodelssurvey,Chowdhery2023palm}, code generation~\cite{chen2021codex}, and dialogue-based assistance~\cite{Murogaki2025llm_dialogue}. As these models are increasingly deployed as autonomous agents, they are expected not only to generate fluent and contextually appropriate responses, but also to adhere to task-specific requirements such as business rules, domain constraints, safety policies, and logical consistency~\cite{shi2024llm_safety,bai2022constitutionalaiharmlessnessai}. 
In practice, correctness therefore depends not only on linguistic plausibility but also on satisfying structured specifications and implicit constraints, including domain rules and user preferences.

Despite their impressive generative abilities, LLM-based agents frequently deviate from such specifications. Common failure modes include hallucinations, constraint violations, incorrect tool usage, flawed reasoning steps, and inconsistent decision-making~\cite{ji2023hallucination,perez2022redteaminglanguagemodels,maynez2020llm_faithfulness}. These issues have been extensively documented in both standalone generation and tool-augmented agent settings~\cite{yao2023ReAct,qin2024toollearningfoundationmodels}. In many real-world deployments, particularly customer-facing or operational settings, such failures are costly because the system often has only a single opportunity to act. Iterative retries or post-hoc corrections may be impractical, making \textit{first-attempt correctness} a central challenge~\cite{weidinger2021ethicalsocialrisksharm,bommasani2022opportunitiesrisksfoundationmodels}.

Several research directions aim to improve the reliability of LLM agents. Reflection-based methods~\cite{shinn2023reflexion,yao2024retroformer} allow models to critique and revise their own outputs, while approaches using LLMs as judges~\cite{zheng2023llm_judge,liu2023G_eval} or debate-style reasoning~\cite{khan2024llm_debate} introduce additional evaluation or deliberation stages. These techniques can significantly improve performance, but they typically rely on repeated attempts or post-hoc refinement. As a result, they are less suitable for interactive applications where agents must succeed within a single execution.

Another line of work adopts actor–critic frameworks inspired by reinforcement learning~\cite{zhang2024LLaMAC_control_llm,dong2025lac_decision_llm_ac,estornell2025acc_collab,yang2025critic_guided_improvement}, in which a critic evaluates the actor’s decisions to guide policy improvement. Recent LLM-based extensions typically assume a \textit{symmetric actor–critic} setting, where both actor and critic are open-source and jointly trainable models, often sharing the same architecture. While this symmetry enables coordinated optimization, it limits applicability in practical deployments. In many real-world systems, the most capable LLMs are proprietary or accessible only through APIs~\cite{openai2024gpt4}, making actor fine-tuning infeasible.

Moreover, existing actor–critic approaches are commonly evaluated in environments with structured action spaces, such as ALFWorld~\cite{shridhar2021alfworld} or WebShop~\cite{yao2023webshop}, where the actor proposes a small set of candidate actions that the critic can rank or score. This assumption does not hold in conversational settings, where actions consist of open-ended language and interactions evolve over multiple turns. Consequently, existing symmetric actor–critic methods are difficult to apply to user-facing LLM agents that rely on proprietary actors and must operate under open-ended conversational dynamics.

These limitations reveal a fundamental tension in existing approaches. Reflection-based methods and LLM-as-a-judge frameworks can leverage powerful proprietary LLMs but offer limited control over their behavior and often rely on retries to recover from errors. In contrast, fine-tuning-based \textit{symmetric actor–critic} frameworks provide stronger control but assume trainable actors and structured action spaces, making them difficult to apply when the actor is proprietary or when actions consist of free-form language in long conversational interactions. This raises a central question:

\noindent\fbox{%
\parbox{\linewidth}{%
\textit{Can we design an agentic system that leverages the capabilities of proprietary LLMs while retaining the adaptability and controllability of open-source models, without relying on retries, actor fine-tuning, or structured action spaces?}
}%
}

To address this challenge, we propose an \textit{asymmetric actor–critic} framework with runtime supervision, as shown in Figure~\ref{fig:teaser}. The system consists of two heterogeneous components:
(1) a primary \textit{actor}, a powerful proprietary LLM responsible for dialogue generation, tool usage, and decision-making; and
(2) a secondary \textit{critic}, a lightweight open-source model that monitors the actor and intervenes during execution when necessary.

Unlike symmetric actor–critic methods that jointly train actors and critics or rank candidate actions, our framework supervises a fixed actor operating in an open-ended action space. The critic provides \textit{runtime supervision} within a single execution trajectory, detecting potential errors, such as hallucinations, policy violations, or inconsistent reasoning, and intervening before the actor finalizes its response.

This design exploits a natural asymmetry between generation and verification. While high-quality generation typically requires large, expressive models, oversight and error detection can often be performed effectively by smaller models~\cite{song2025mindgap_gen_veri, zhou2025ICLR}. By separating these roles, our framework combines the capabilities of frontier proprietary LLMs with the flexibility and adaptability of open-source models, enabling reliable and controllable behavior in one-shot, user-facing deployments.

We evaluate the proposed framework in customer service and travel planning environments using $\tau$-bench~\cite{yao2024tau_bench} and UserBench~\cite{qian2025userbench}. These benchmarks require agents to operate under rich domain-specific specifications, including company policies, procedural constraints, and evolving user preferences, and emphasize success within a single execution trajectory. Across both benchmarks, introducing a runtime critic substantially improves reliability, reducing policy violations and hallucinations while achieving higher task success rates compared to strong single-actor baselines. Notably, we find that relatively small open-source critics based on Qwen3 models with up to 32B parameters~\cite{yang2025qwen3technicalreport} perform comparably or even better than a much larger proprietary critic, Claude-4 Sonnet~\cite{anthropic2024claude4}, when supervising the same actor, highlighting the generation-verification gap.

To further demonstrate the domain adaptability of our framework, we introduce a self-play, critic-in-the-loop data generation pipeline for training critics in asymmetric actor–critic settings. 
Unlike prior approaches that focus on improving the actor’s generative behavior, our pipeline focuses on learning supervisory signals for critic interventions.
By repurposing existing open-source conversational datasets, the pipeline constructs supervision signals for critic intervention decisions, enabling efficient supervised fine-tuning without modifying the proprietary actor.
Overall, our results show that effective agent control can emerge from asymmetric cooperation between a fixed high-capability actor and a lightweight, trainable critic.

\bigskip
\noindent
\textbf{Our contributions} are summarized as follows:

\begin{enumerate}[topsep=0mm, itemsep=0mm, leftmargin=*]
    \item \textbf{Asymmetric Actor–Critic Runtime Supervision.}
    We introduce an asymmetric actor–critic framework that enables runtime supervision of fixed, proprietary LLM actors. A lightweight critic monitors and intervenes within a single execution trajectory, providing control over open-ended language actions without actor retraining, candidate action enumeration, or retries.

    \item \textbf{Critic-Oriented Data Generation Pipeline.}
    We propose a data generation pipeline that repurposes existing conversational datasets to produce supervision signals for critic intervention decisions, enabling efficient supervised fine-tuning of the critic while leaving the proprietary actor unchanged.
    
    \item \textbf{Empirical Validation in Conversational Agent Benchmarks.}
    We evaluate the framework on $\tau$-bench and UserBench, showing consistent improvements in reliability and task success over strong single-actor and state-of-the-art baselines in realistic multi-turn conversational settings.
\end{enumerate}

\section{Related Work}

\textbf{Symmetric Actor–Critic LLM Agent Training.}
The actor–critic paradigm originates from reinforcement learning~\cite{mnih2016a3c,lillicrap2019a2c}, where an actor proposes actions while a critic evaluates them to guide policy updates. Recent work extends this paradigm to LLM-based agents, typically assuming that both actor and critic are open-source and jointly trainable models. 

Several approaches follow this formulation. LLaMAC~\cite{zhang2024LLaMAC_control_llm} proposes an actor–critic architecture for multi-agent decision-making with LLMs. Dong et al.~\cite{dong2025lac_decision_llm_ac} introduce long-horizon Q-like evaluations to guide policy improvement, while Hong et al.~\cite{hong2025llm_ac} represent policies and value functions in natural language for planning tasks. ACC-Collab~\cite{estornell2025acc_collab} trains actor–critic agents collaboratively using preference-based optimization, and CGI~\cite{yang2025critic_guided_improvement} trains a critic to generate feedback that the actor incorporates through iterative refinement. 

These methods focus on \emph{policy improvement through training} and require updating the actor. In contrast, our work considers deployment settings where the actor is fixed and potentially proprietary. Moreover, existing actor–critic systems typically operate in environments with structured or enumerable action spaces, such as ALFWorld~\cite{shridhar2021alfworld} and WebShop~\cite{yao2023webshop}, where candidate actions are evaluated or ranked. Such assumptions break down in long conversational tasks, where actions correspond to free-form language responses.

\textbf{Prompt and Policy Optimization Approaches.}
Another line of work improves agent behavior by optimizing prompts or reasoning structures without modifying model parameters. For example, PACE~\cite{dong2024pace} treats prompts as policies and applies actor–critic optimization to iteratively edit them, while RePrompt~\cite{chen2025reprompt} refines prompts for multi-step agents using intermediate feedback. These approaches show that prompt-level optimization can improve agent performance, but they are mainly designed for offline or iterative improvement rather than runtime supervision in one-shot execution settings.

\textbf{Reflection and Retrospective Models.}
Reflection-based methods improve agent behavior by analyzing errors and generating corrective reasoning signals. Reflexion~\cite{shinn2023reflexion} enables agents to learn from past failures by storing reflective feedback that guides future attempts. Retroformer~\cite{yao2024retroformer} trains a retrospective model to revise prompts based on environment feedback, with subsequent work improving data efficiency and stability~\cite{feng2025retrospective_lm}. CRITIC~\cite{gou2024CRITIC} introduces a critique-in-the-loop framework where the LLM uses tools to diagnose errors and refine its outputs during execution. 
These approaches share our goal of improving agent reliability through auxiliary critique mechanisms. However, most reflection-based methods, including Reflexion and Retroformer, primarily improve behavior across multiple attempts, whereas our framework focuses on influencing the actor’s decisions within a single execution trajectory.

\textbf{LLMs as Judges and Runtime Overseers.}
Recent work also explores using LLMs as evaluators or runtime monitors for agent behavior. LLM-as-a-judge frameworks assess outputs or trajectories to provide feedback or ranking signals for model improvement~\cite{zheng2023llm_judge, liu2023G_eval, khan2024llm_debate}. Related runtime oversight systems introduce supervisory components that monitor agent behavior during execution. For example, Task Shield~\cite{jia2024task_shield} checks whether agent actions remain aligned with user goals, mitigating prompt injection and task hijacking. Pro2Guard~\cite{wang2026pro2guard} predicts undesirable behaviors from agent traces, while ShieldAgent~\cite{chen2025shieldagent} enforces safety constraints through reasoning-based verification. These systems share our runtime monitoring perspective but focus primarily on safety and security guarantees. In contrast, our work addresses general agentic control under domain-specific specifications such as policies, procedural rules, and user preferences.

\section{Problem Statement}

\textbf{Multi-turn Conversational Decision Process.}
We model the interaction between an agent and a user as a multi-turn conversational decision process over a horizon $T$. 
At each turn $t \in [T]$, the agent observes the interaction history
$h_t = (o_1, a_1, o_2, a_2, \dots, o_t),
$
where $o_t$ denotes the observation at turn $t$, consisting of the user utterance and any environment feedback (e.g., tool outputs), and $a_t$ denotes the agent's action.
Actions may include natural language responses, structured tool calls, or other task-specific operations.

Given the history, the agent selects an action $a_t$, which is executed in the environment and produces the next observation $o_{t+1}$. The interaction proceeds until termination, producing a trajectory
\[
\vspace{-4pt}
\tau = (o_1, a_1, o_2, a_2, \dots, o_T, a_T),
\]

A task-specific evaluation function $R(\tau) \in [0, 1]$ measures the quality of the trajectory, for example by evaluating whether the task meets certain specifications and succeeds.

\section{Asymmetric Actor--Critic Framework}
\label{sec:asymmetric_ac}
We instantiate the agent using an asymmetric actor–critic (\asymac) architecture consisting of two components (see Appendix~\ref{subsec:appendix_ac_algorithm} for the full pseudocode):

\begin{itemize}[itemsep=0mm, topsep=0mm, leftmargin=*]
    \item Actor $\pi_{A}$: a high-capability proprietary LLM with fixed parameters responsible for generating actions. 
    \item Critic $C_{\theta}$: 
    a smaller open-source model with trainable parameters $\theta$ that supervises the actor and intervenes during execution when necessary.
\end{itemize}

\subsection{Runtime Critic Supervision}
At each turn $t$, the actor first proposes a candidate action $\widehat{a}_t \sim \pi_A(\cdot \mid h_t)$.

Let $\delta_t \in \{0,1\}$ denote whether the critic is invoked to supervise the action at turn $t$.  
If $\delta_t = 1$, the critic evaluates the proposal and produces verbal feedback $c_t = C_\theta(h_t, \widehat{a}_t)$,
which may indicate approval, request revision, highlight missing information, or provide corrective guidance.  
The feedback is appended to the context, and the actor generates a revised action $a_t \sim \pi_A(\cdot \mid h_t, \widehat{a}_t, c_t)$.

If $\delta_t = 0$, the critic is not invoked and the proposal is executed directly, i.e., $a_t = \widehat{a}_t$.
The final action $a_t$ is then executed in the environment.

\subsection{Learning Objective}
Recall that the actor remains fixed in our framework.
Training therefore focuses solely on the critic, which learns to intervene when necessary to improve trajectory outcomes.
The objective is
\[
\vspace{-5pt}
\max_{\theta} \; \mathbb{E}_{\tau \sim (\pi_A, C_\theta)}[R(\tau)].
\]
where trajectories are generated by the actor under critic supervision.

\subsection{Advantages of Asymmetric Actor--Critic}

The asymmetric design separates generation from verification. A powerful proprietary actor provides strong language and reasoning capabilities, while a lightweight open-source critic performs oversight without modifying the actor. This enables practical customization in deployment settings where the actor is fixed or inaccessible. Because the critic performs evaluation rather than generation, it can be significantly smaller than the actor, making adaptation efficient and low-cost while still benefiting from the generation–verification gap~\cite{song2025mindgap_gen_veri, zhou2025ICLR}. To support task-specific customization, we introduce in the next section a self-play data generation pipeline that produces supervision signals for training the critic.

\section{Data Generation for Critic Training}
\label{sec:critic_customization}

Most existing conversational datasets are designed for actor training. In supervised fine-tuning (SFT), a model learns to generate final responses that directly satisfy user requests. However, such datasets do not provide supervision for a critic, whose role is not to produce responses but to evaluate and guide the actor’s behavior during interaction.

To address this gap, we introduce a critic-oriented data generation pipeline that repurposes existing conversational datasets to produce supervision signals for critic training. Our pipeline converts single-actor trajectories into critic supervision signals and applies trajectory filtering to ensure that the resulting SFT data captures informative intervention opportunities in multi-turn conversations.

Unlike most existing data generation pipelines, which rely on a proprietary model to produce expert demonstrations that a trainable model then imitates, e.g.,~\cite{prabhakar2025apigenmt, yang2025critic_guided_improvement}, our method adopts a self-play, critic-in-the-loop data curation paradigm. The critic interacts with the actor, evaluates its behavior online, and generates supervision signals from its own intervention decisions.
The data generation pipeline is detailed below, with the full procedure presented in Algorithm~\ref{alg:critic_data_generation} in Appendix~\ref{subsec:appendix_data_gen_algorithm}.

\subsection{Data Preparation}

Let $\mathcal{D} = \{\tau^{(i)}\}_{i=1}^{N}$ denote an existing multi-turn conversational dataset, where each trajectory
\vspace{-4pt}
\[
\tau^{(i)} = (o_1^{(i)}, a_1^{(i)}, \dots, o_{T^{(i)}}^{(i)}, a_{T^{(i)}}^{(i)})
\vspace{-4pt}
\]
represents a successful dialogue between a user and an AI assistant. Here $o_t^{(i)}$ denotes the observation at turn $t$ (e.g., user utterances and environment feedback) and $a_t^{(i)}$ denotes the assistant action.

Our goal is not to imitate the assistant that generated these trajectories, but to train a critic that can effectively supervise a fixed actor model $\pi_A$. 
Because the assistant in the dataset may differ from $\pi_A$, the original trajectories cannot be used directly. Instead, we treat each trajectory as an implicit demonstration of an underlying task, reconstruct the task specification, and replay it using the target actor $\pi_A$.

Specifically, for each trajectory $\tau^{(i)}$, we infer a latent task specification $x^{(i)}$ using a language model, capturing the user's intent, task objective, and relevant constraints. We also derive a task success criterion $R(\tau) \in [0,1]$ from the original trajectory—for example, whether the required actions are correctly executed—which will later be used to evaluate replayed interactions.

Replaying tasks with the target actor $\pi_A$ allows us to identify tasks on which $\pi_A$ struggles; these tasks are later used to generate critic supervision signals through actor–critic interactions. In contrast, tasks that $\pi_A$ already solves reliably provide little useful signal for training the critic. Consequently, the inferred task specifications $x^{(i)}$ form the basis for the subsequent task filtering and trajectory generation steps.

\vspace{-3pt}
\subsection{Task Filtering}

To focus training on cases where critic intervention is most beneficial, we identify a subset of \textit{hard tasks} $\mathcal{H}$ on which a single actor frequently fails. Critic supervision is most informative when the actor alone struggles but the actor–critic system can succeed through intervention.

For each task $x^{(i)}$, $i \in [N]$, we execute the actor-only baseline for $K$ independent runs, producing trajectories $\{\widehat{\tau}_k^{(i)}\}_{k=1}^{K}$. A task is classified as hard if the number of failed runs exceeds a threshold $\psi$:
\vspace{-8pt}
\[
\sum_{k=1}^{K} \mathbb{I}[R(\widehat{\tau}_k^{(i)}) < 1] \ge \psi,
\vspace{-8pt}
\]
where $R(\widehat{\tau}_k^{(i)}) < 1$ indicates failure on task $x^{(i)}$. The set of such tasks forms the hard task set $\mathcal{H}$, which is retained for critic data generation.

\subsection{Actor--Critic Trajectory Collection}
For each hard task $x \in \mathcal{H}$, we execute the asymmetric actor–critic framework $(\pi_A, C_\theta)$ also for $K$ runs. 
A replayed trajectory by actor-critic $\widetilde{\tau}$ is retained if it satisfies:
\begin{enumerate}[topsep=0pt, itemsep=0pt, parsep=0pt, partopsep=0pt]
    \item Task success: $R(\widetilde{\tau}) = 1$.
    \item Non-trivial intervention: There exists a turn $t$ in the trajectory $\widetilde{\tau}$ such that the verbal feedback by the critic $c_t$ rejects or revises the proposed action by the actor $a_t$. 
\end{enumerate}
The second condition ensures that retained trajectories contain meaningful critic interventions rather than only passive approvals.

\subsection{Training Sample Extraction}

Each retained trajectory $\widetilde{\tau}$ is segmented into steps involving critic evaluation.
For every time step $t$ where the critic evaluates an actor proposal, we construct a supervised training example consisting of a prompt $x_t$ and a completion sentence label $y_t$:
\vspace{-6pt}
\[
x_t = (h_t, a_t), \quad y_t = c_t
\vspace{-3pt}
\]
where $h_t$ is the dialogue context up to turn $t$, $a_t$ is the actor’s proposed action, and 
$c_t$ is the critic’s evaluation (e.g., approve, reject, or corrective guidance).
The resulting supervised dataset is:
\[
\vspace{-4pt}
\mathcal{S} = \cup_{\widetilde{\tau}} \{(x_t, y_t)\}_{(x_t,y_t) \in \widetilde{\tau}}
\]
which is used to fine-tune the critic $C_\theta$
via supervised fine-tuning. A single trajectory may yield multiple training samples.

\section{Experiments}

In this section, we evaluate \asymac\ in three stages. First, we assess the effectiveness of the framework without critic training by comparing it to a single-actor baseline. Second, we vary the critic model size while fixing the proprietary actor to examine the impact of critic capacity. Third, we fine-tune the critic using the proposed synthetic data generation pipeline and evaluate the resulting performance gains against state-of-the-art baselines, validating both critic customization and the effectiveness of our data generation approach.

\subsection{Datasets and Metrics}

\textbf{Datasets.} We evaluate on two representative multi-turn conversational benchmarks (see Appendix~\ref{sec:appendix_benchmarks} for more details). The number of test samples for evaluation is reported in Table~\ref{tab:datasets_summary}.
\begin{enumerate}[itemsep=0pt, topsep=0mm, parsep=0pt, partopsep=0pt, leftmargin=*]
    \item \textbf{$\tau$-bench}~\cite{yao2024tau_bench}: a customer service agent interacts with a user to fulfill requests such as canceling flights, modifying bookings, or returning delivered products. Tasks involve policy compliance, tool use, and multi-step reasoning under domain constraints. Success is determined by whether the final system state matches a ground truth system state.

    \item \textbf{UserBench}~\cite{qian2025userbench}: a travel planning agent must recommend exactly one option for each travel aspect (e.g., apartment, restaurant, rental car) from a structured database. During a multi-turn interaction, the agent must proactively ask questions and infer both explicit and implicit user preferences. An option is considered optimal if it satisfies all user preferences while minimizing cost among the available ones.
\end{enumerate}

For $\tau$-bench, we evaluate in the retail and airline domains. 
For UserBench, we evaluate in the widely used \texttt{Travel-22} (T-22), \texttt{Travel-33} (T-33), and \texttt{Travel-44} (T-44) environments. In this notation, \texttt{Travel-$xx$} (T-$xx$) denotes tasks involving two travel aspects, where each aspect contains $x$ implicit user preferences that must be identified and satisfied. These configurations represent increasing levels of difficulty, with T-22 being the easiest and T-44 the most challenging due to the larger number of preferences per aspect.
Our evaluation focuses on the single-choice setting, where the actor is restricted to a single recommendation per travel aspect.
\vspace{-5pt}

\begin{table}[h]
    \centering
     \caption{A summary of datasets statistics.}
    \begin{adjustbox}{width=0.35\textwidth}
    \begin{tabular}{|c|c|c|c|}
    \hline
        \multicolumn{2}{|c|}{Dataset} & \# train (for critic) & \# test \\
    \hline
        \multirow{2}{*}{$\tau$-bench} & airline & 2363 & 50 \\
        & retail & 1184 & 115 \\
    \hline
        \multirow{3}{*}{UserBench} & T-22 & 570 & 101 \\
        & T-33 & 489 & 87 \\
        & T-44 & 379 & 67 \\
    \hline
    \end{tabular}
    \end{adjustbox}
    \vspace{-10pt}\label{tab:datasets_summary}
\end{table}

\textbf{Evaluation Metrics.}
For $\tau$-bench, the reward is binary, $R(\tau) \in \{0, 1\}$. A trajectory $\tau$ receives $R(\tau) = 1$ if the final system state matches the ground truth, and 0 otherwise. We report the \textit{Task Success Rate} $\Pr[R(\tau) = 1]$ (pass@1), averaged over five independent runs per task.

For UserBench, $R(\tau) \in \{0, 0.8, 1\}$. For each travel aspect, selecting the optimal option yields $R(\tau)=1$, selecting a correct but non-optimal option yields $R(\tau) = 0.8$, and selecting a wrong option yields $R(\tau) = 0$. Following~\cite{qian2025userbench}, we report \textit{score} as the primary evaluation metric, defined as the average reward $R(\tau)$ across all travel aspects and five independent runs.

\subsection{Effectiveness of Asymmetric Actor-Critic}

Our first goal is to validate the effectiveness of the proposed asymmetric actor-critic without training the critic, by comparing it against a single proprietary actor. 
We use Claude-4 Sonnet as the actor model and Qwen3-8B as the critic model. 

\subsubsection{Implementation Details}

\textbf{Role of the Critic as a Constraint Verifier.}
In both $\tau$-bench and UserBench, the critic is implemented as a domain-specific constraint verification module that supervises the actor’s key decisions during interaction (see the full prompt used in our implementation in Appendix~\ref{subsec:appendix_ac_prompts}).

In $\tau$-bench, constraints are explicitly defined by written policies (e.g., conditions under which a flight may be canceled or a return request processed). The critic verifies that the actor’s actions comply with these policies.

In UserBench, constraints arise implicitly from user preferences expressed during conversation (e.g., preferring a restaurant with rating $\geq 7$ while minimizing price). The critic ensures that recommendations satisfy all stated or inferred preferences and prompts the actor to query the user when critical preferences remain unspecified.

When invoked, the critic evaluates two types of errors:
\begin{enumerate}[topsep=0pt,itemsep=0pt,parsep=0pt,partopsep=0pt,leftmargin=*]
\item \textit{Constraint violation}. The actor permits an action that violates domain policies or user preferences.
\item \textit{Constraint hallucination}. The actor incorrectly blocks a valid action due to hallucinated or misapplied constraints (e.g., assuming nonexistent user preferences in UserBench).
\end{enumerate}

Recall that $\delta_t \in \{0,1\}$ indicates whether the critic intervenes by providing feedback to the actor.
In $\tau$-bench, $\delta_t=1$ when the actor proposes an action that changes the system state, such as issuing tool calls to book, cancel, or modify a flight.
In UserBench, $\delta_t=1$ occurs when the actor proposes a final recommendation for a travel aspect.

\begin{table}[h]
    \centering
    \caption{Task success rate (pass@1) of single actor and \asymac\ comparison on $\tau$-bench.}
    \begin{adjustbox}{width=0.4\textwidth}
    \begin{tabular}{|c|c|c|c|}
    \hline
        Method & Retail & Airline & Average \\
    \hline
        Single Actor & 0.6956 & 0.5040 & 0.5998 \\
    \hline
        \asymac & 0.7356 & 0.5280 & 0.6318 \\
    \hline
    \end{tabular}
    \end{adjustbox}
    \label{tab:asym_ac_tau_bench}
    \vspace{-15pt}
\end{table}

\begin{table}[h]
    \centering
    \caption{Score of single actor and \asymac\ comparison on UserBench.}
    \begin{adjustbox}{width=0.48\textwidth}
    \begin{tabular}{|c|c|c|c|c|}
    \hline
        Method & T-22 & T-33 & T-44 & Average \\
    \hline
        Single Actor & 0.3780 & 0.3143 & 0.2907 & 0.3277 \\
    \hline
        \asymac & 0.4356 & 0.3566 & 0.3472 & 0.3798 \\
    \hline
    \end{tabular}
    \end{adjustbox}
    \vspace{-15pt}
    \label{tab:asym_ac_userbench_best}
\end{table}

\subsubsection{Results and Discussion}

The results on $\tau$-bench and UserBench are reported in Table~\ref{tab:asym_ac_tau_bench} and~\ref{tab:asym_ac_userbench_best}. Across both benchmarks and all evaluated domains and environments, \asymac\ consistently outperforms the single-actor baseline using Claude-4 Sonnet alone, demonstrating the effectiveness of the \asymac\ in improving performance on multi-turn conversational tasks.

\subsection{Critic Capacity}

Our next goal is to investigate how the capacity of the critic model influences the performance of \asymac. Specifically, we examine whether a lightweight critic suffices for effective supervision.

\textbf{Comparison.} To ensure a fair comparison, we fix the actor (A) across all settings to Claude-4 Sonnet and vary only the critic model. We evaluate \asymac\ using four critic (C) models from different model families and of varying sizes: Claude-4 Sonnet and Qwen3-\{4, 8, 32\}B. The performance of single actor serves as a reference point.

\begin{table}[]
    \centering
    \caption{Task success rate (pass@1) on $\tau$-bench of single actor and \asymac\ with different models. }
    \begin{adjustbox}{width=0.48\textwidth}
    \begin{tabular}{|c|c|c|c|c|}
    \hline
        Method & Models & Retail & Airline & Average \\
    \hline
        Actor Only & Claude-4(A) & 0.6956 & 0.5040 & 0.5998 \\
    \hline
        \asymac\ & Claude-4(A) + Claude-4(C) & 0.7270 & 0.4920 & 0.6095 \\
        \asymac\ & Claude-4(A) + Qwen3-4B(C) & 0.7217 & 0.5280 & 0.6249 \\
        \asymac\ & Claude-4(A) + Qwen3-8B(C) & 0.7356 & 0.5280 & 0.6318 \\
        \asymac\ & Claude-4(A) + Qwen3-32B(C) & 0.7652 & 0.5400 & 0.6526 \\
    \hline
    \end{tabular}
    \end{adjustbox}
    \label{tab:results_tau_bench_critic_capacity}
\end{table}

\begin{table}[]
    \centering
    \caption{Score on UserBench of single actor and \asymac\ with different models. }
    \begin{adjustbox}{width=0.48\textwidth}
    \begin{tabular}{|c|c|c|c|c|c|}
    \hline
        Method & Models & T-22 & T-33 & T-44 & Average \\
    \hline
        Actor Only & Claude-4(A) & 0.3780 & 0.3143 & 0.2907 & 0.3277 \\
    \hline
        \asymac\ & Claude-4(A) + Claude-4(C) & 0.4046 & 0.3384 & 0.2860 & 0.3430 \\
        \asymac\ & Claude-4(A) + Qwen3-4B(C) & 0.4259 & 0.3197 & 0.3224 & 0.3560 \\
        \asymac\ & Claude-4(A) + Qwen3-8B(C) & 0.4356  & 0.3566 & 0.3472 & 0.3798 \\
        \asymac\ & Claude-4(A) + Qwen3-32B(C) & 0.4574 & 0.3611 & 0.3466 & 0.3884 \\
    \hline
    \end{tabular}
    \end{adjustbox}
    \vspace{-15pt}\label{tab:results_userbench_critic_capacity_best}
\end{table}

\textbf{Results and Discussion.}
Results on $\tau$-bench and UserBench are reported in Table~\ref{tab:results_tau_bench_critic_capacity} and Table~\ref{tab:results_userbench_critic_capacity_best}.
When comparing critic capacity, we observe that using Claude-4 Sonnet as the critic consistently underperforms Qwen3-8B and Qwen3-32B across both benchmarks. This suggests that a highly capable proprietary model is not necessary for the critic role: strong generative ability does not automatically translate into better supervision, even without critic fine-tuning. In contrast, within the same model family, larger critics tend to perform better. For example, in the Qwen3 series, performance improves as model size increases.

\subsection{Customization through Critic Fine-tuning}

Given the consistent gains of \asymac\ over the single-actor baseline, we next investigate whether further improvements can be achieved by customizing the critic via supervised fine-tuning (SFT), denoted as \asymacsft. In this section, we evaluate \asymacsft\ with a fine-tuned critic and compare it against strong state-of-the-art baselines.

\subsubsection{Implementation Details}

\textbf{Sources of SFT Data for the Critic.}
To customize the critic via supervised fine-tuning, we leverage different data sources depending on the benchmark. $\tau$-bench does not provide unified training data across its domains. We therefore use the open-sourced synthetic datasets generated by the pipeline proposed in~\cite{prabhakar2025apigenmt}. Given the substantial differences between the airline and retail domains, we train separate critics for each domain using domain-specific synthetic data.

UserBench provides explicit training and test splits. We train the critic using the training data and evaluate on the corresponding test data. As the T-22, T-33 and T-44 environments belong to the same domain, we train a single critic using data aggregated across all three environments and report results separately for each.

In generating training samples for the critic, we set the number of repeated runs to $K=5$ and the hard-task threshold to $\psi=2$. 
The number of generated critic training samples across benchmarks is summarized in Table~\ref{tab:datasets_summary}; additional dataset statistics are provided in Appendix~\ref{subsec:appendix_add_statistics}.

\textbf{Critic Training Details.} 
We fine-tune the critic using the TRL framework, with Qwen3-8B as the base model and LoRA~\cite{hu2021lora} for parameter-efficient adaptation. We set the LoRA rank to 16, $\alpha=32$, and dropout 0.1. Optimization is performed with AdamW (default settings), a warmup ratio of 0.08, and gradient accumulation of 16 steps. We search learning rates in \{$10^{-3}, 10^{-4}, ..., 10^{-6}$\}, and train for 5–15 epochs.

\subsubsection{Baselines}

We compare against the following baselines:

\begin{enumerate}[itemsep=0pt, topsep=0mm, parsep=0pt, partopsep=0pt, leftmargin=*]

\item \textbf{\asymac\ (Section~\ref{sec:asymmetric_ac}).} 
The proposed asymmetric actor–critic framework with an off-the-shelf critic and no task-specific fine-tuning.

\item \textbf{CRITIC}~\cite{gou2024CRITIC}. 
A single-actor self-critique method using in-context critique demonstrations. We adapt CRITIC to our one-shot conversational settings (see Appendix~\ref{sec:appendix_baselines}).

\item \textbf{ReAct}~\cite{yao2023ReAct}. 
A reasoning-and-acting framework that alternates between reasoning traces and actions. On $\tau$-bench we use the native ReAct implementation; on UserBench the same interaction pattern is enforced by the environment interface.

\item \textbf{APIGen-MT}~\cite{prabhakar2025apigenmt}. 
A data-centric approach that generates synthetic multi-turn conversations to fine-tune open-source agents. Evaluated only on $\tau$-bench; we report results from the best-performing model, \textit{xLAM-2-70b-fc-r}.

\item \textbf{IRMA}~\cite{Mishra2025IRMA}. 
An input reformulation approach that improves robustness to policy violations by rewriting user inputs. This baseline is $\tau$-bench specific.
\end{enumerate}

Across all methods, we use Claude-4 Sonnet as the actor model. In the \asymacsft, we additionally use Qwen3-8B as the critic.
We adopt the canonical baseline implementations provided by each benchmark. On UserBench, the native agent interface follows a ReAct-style~\cite{yao2023ReAct} interaction loop, so all methods naturally operate in this paradigm and we do not report ReAct separately as a baseline.

\begin{table}[]
    \centering
    \caption{Task success rate (pass@1) on $\tau$-bench across different methods and domains. }
    \begin{adjustbox}{width=0.4\textwidth}
    \begin{threeparttable}
    \begin{tabular}{|c|c|c|c|}
    \hline
        Method & Retail & Airline & Average \\
    \hline
        \asymac\ (ours) & 0.7356 & 0.5280 & 0.6318 \\
        CRITIC & 0.7061 & 0.5200 & 0.6131 \\
        ReAct & 0.7130  & 0.5040 & 0.6085 \\
        \textit{xLAM-2-70b-fc-r} $\dagger$
        & 0.671 & 0.452 & 0.562 \\
        IRMA & 0.583 & 0.472 & 0.5275 \\
    \hline
        \asymacsft\ (ours) & \textbf{0.7652} & \textbf{0.5440} & \textbf{0.6546} \\
    \hline
    \end{tabular}
    \begin{tablenotes}
    \footnotesize
    \item $\dagger$: Best-performing open-source model by \textbf{APIGen-MT}
\end{tablenotes}

    \end{threeparttable}
    \end{adjustbox}
    \vspace{-3pt}
    \label{tab:results_tau_bench}
\end{table}

\begin{table}[]
    \centering
    \caption{Score on UserBench across different methods and environments. }
    \begin{adjustbox}{width=0.48\textwidth}
    \begin{tabular}{|c|c|c|c|c|}
    \hline
        Method & T-22 & T-33 & T-44 & Average \\
    \hline
        \asymac\ (ours) & 0.4356 & 0.3566 & 0.3472 & 0.3798 \\
        CRITIC & 0.4332 & 0.3455 & 0.3069 & 0.3619 \\
        % ReAct & & & & \\
        % LLM-as-Judge & & & & \\
    \hline
        \asymacsft\ (ours) & \textbf{0.4653} & \textbf{0.4161} & \textbf{0.3713} & \textbf{0.4176} \\
    \hline
    \end{tabular}
    \end{adjustbox}
    \vspace{-15pt}\label{tab:results_userbench_best}
\end{table}

\subsubsection{Results and Discussion} 
Results on $\tau$-bench and UserBench are reported in Tables~\ref{tab:results_tau_bench} and~\ref{tab:results_userbench_best}. 
Across both benchmarks, \asymacsft\ consistently outperforms \asymac\ with an untrained critic, further improving over the single-actor baseline and all other comparison methods. 
These results demonstrate that supervised fine-tuning of the critic using the proposed data generation pipeline provides additional performance gains. 
We present qualitative examples in Appendix~\ref{subsect:appendix_examples} illustrating how critic interventions correct actor errors across different domains.

\section{Conclusion}

We propose an asymmetric actor–critic framework for multi-turn conversational agents, where a fixed proprietary actor is supervised by a customizable open-source critic. Experiments on $\tau$-bench and UserBench show consistent improvements over strong single-actor baselines. Our results demonstrate that effective supervision does not require critics matching the actor’s scale: lightweight critics can provide substantial gains. We further introduce a synthetic data generation pipeline for critic fine-tuning, enabling efficient customization without modifying the proprietary actor. These findings highlight asymmetric actor–critic as a practical approach for improving the reliability and controllability of real-world LLM systems.
Broader impact and potential risks are discussed in Appendix~\ref{sec:appendix_risks_broader_impact}.

\clearpage
\section*{Limitations}
While the proposed asymmetric actor--critic framework improves reliability in multi-turn conversational agents, several limitations remain. First, the effectiveness of the approach still depends on the underlying actor model’s capability; if the actor lacks sufficient reasoning ability, critic feedback may not fully correct its behavior. Second, runtime supervision introduces additional computational cost and latency due to the critic evaluation step.

% Custom bibliography entries only
\bibliography{mybib}

\clearpage
\appendix

\section{Asymmetric Actor--Critic}
\label{sec:appendix_ac}

\subsection{Algorithm}
\label{subsec:appendix_ac_algorithm}

We present the pseudocode of the asymmetric actor--critic framework in Algorithm~\ref{alg:asym_ac_inference}.

\begin{algorithm}[H]
\caption{Asymmetric Actor--Critic}
\label{alg:asym_ac_inference}
\begin{algorithmic}[1]
\REQUIRE Fixed actor $\pi_A$, critic $C_\theta$, initial observation $o_1$, horizon $T$
\STATE Initialize interaction history $h_1 \gets (o_1)$
\FOR{$t = 1$ to $T$}
    \STATE Actor proposes action $\widehat{a}_t \sim \pi_A(\cdot \mid h_t)$
    \STATE $\delta_t \gets \textsc{ShouldIntervene}(\widehat{a}_t)$
    \IF{$\delta_t = 1$}
        \STATE Critic generates feedback: 
        
        $c_t \gets C_\theta(h_t, a_t)$
        \STATE Actor revises action: 
        
        $a_t \gets \pi_A(\cdot \mid h_t, \widehat{a}_t, c_t)$
    \ELSE
        \STATE $a_t \gets \widehat{a}_t$
    \ENDIF
    \STATE Actor executes $a_t$ and observe $o_{t+1}$
    \STATE Update history $h_{t+1} \gets (h_t, \tilde{a}_t, o_{t+1})$
    \IF{termination condition is met}
        \STATE \textbf{break}
    \ENDIF
\ENDFOR
\STATE \textbf{return} Full trajectory 

$\tau = h_T=(o_1, a_1,\dots, o_T, a_T)$
\end{algorithmic}
\end{algorithm}

\subsection{Full Prompts}
\label{subsec:appendix_ac_prompts}
The actor’s system and user prompts follow the implementations provided by the benchmarks ($\tau$-bench and UserBench). The system and user prompts used for the critic in the two benchmarks are listed below.

\begin{tcolorbox}[title=Critic System Prompt on $\tau$-bench, breakable, colback=white, colframe=black]
\begin{Verbatim}[breaklines=true, breakanywhere=true]
You are a policy verification Critic Agent. Your role is to evaluate and advise a customer service LLM agent to ensure its actions and responses are fully aligned with domain policies.

# Core Responsibilities

Valid Requests: If the user’s request is valid and complies with all applicable policies, confirm that the customer service agent correctly addresses the request without inventing or misrepresenting policies.

Policy Violations: If the user’s request violates policy, ensure that the customer service agent appropriately declines the request. If the agent fails to do so, advise that it must decline.

# Evaluation Criteria

Compliance Verification: Assess whether the agent’s response and intended actions (including tool calls) comply with all relevant policies.

Argument Validation: Check that the arguments/parameters provided in tool calls also meet policy requirements.

Faithful Execution: Ensure the agent fully executes valid user requests without hallucinating policies or misapplying them.

# Guiding Principle

Always base your judgment on the conversation between the user and the customer service agent, as well as the official policies in force.

The current time is 2024-05-15 15:00:00 EST.
\end{Verbatim}
\end{tcolorbox}

\begin{tcolorbox}[title=Critic User Prompt on $\tau$-bench, breakable, colback=white, colframe=black]
\begin{Verbatim}[breaklines=true, breakanywhere=true]
Below is the information you need to verify whether the agent's proposed action is compliant with the policies.

The Agent's Proposed Action:
<proposed action>

Relevant Domain Policies:
<policies>

Conversation History:
<conversation history>

Your task:
Review the agent's proposed action, the action's arguments (if it contains tool calls), the relevant policy and the conversation history. Determine whether the agent’s action and its arguments (if any) fully comply with the domain policy. 
\end{Verbatim}
\end{tcolorbox}

\begin{tcolorbox}[title=Critic System Prompt on UserBench, breakable, colback=white, colframe=black]
\begin{Verbatim}[breaklines=true, breakanywhere=true]
You are a Preference Verification Critic Agent. Your role is to evaluate and advise a travel planning LLM agent to ensure its recommendations are fully aligned with the user’s expressed or implied preferences.

# Core Responsibilities

Valid Recommendations:
If the agent’s proposed recommendation is consistent with all user preferences revealed in the conversation, confirm that the agent is making a faithful and justified choice without overlooking critical information.

Missing Preference Detection:
If the user’s intent is underspecified or some critical preferences have not been elicited, identify the missing preferences and advise the agent to ask clarifying questions before making a recommendation.

Sub-optimal Recommendations:
If the agent’s proposed recommendation is compatible with the user’s expressed or implied preferences but is not the optimal option (e.g., higher cost, worse value), identify the sub-optimality and explain how the recommendation can be improved.

Redundant Recommendations:
Ensure that the agent recommends at most one option per travel aspect. If the agent attempts to re-recommend or override an existing recommendation, flag this as an error.

# Evaluation Criteria

Preference Coverage:
Assess whether the agent has sufficiently uncovered relevant user preferences associated with the current travel aspect.

Preference Faithfulness:
Verify that the agent’s recommendation faithfully reflects the user’s preferences and does not rely on unsupported assumptions or hallucinated preferences.

# Guiding Principle

Always base your judgment strictly on the explicitly provided options and the user preferences inferred from the conversation. Do not introduce new preference dimensions or external assumptions.
\end{Verbatim}
\end{tcolorbox}

\begin{tcolorbox}[title=Critic User Prompt on UserBench, breakable, colback=white, colframe=black]
\begin{Verbatim}[breaklines=true, breakanywhere=true]
Below is the information you need to verify whether the agent’s proposed recommendation is appropriate.

Travel Aspect Under Consideration:
<travel aspect>

The Agent’s Proposed Recommendation:
<recommendation>

Available Options for This Aspect:
<options>

Conversation History:
<conversation history>

Your task:
Review the agent’s proposed recommendation in light of the available options and the user’s explicitly or implicitly expressed preferences based on the conversation history.

Determine:
1. Whether the proposed recommendation is the optimal choice given the known user preferences.
2. Whether there are critical user preferences relevant to this travel aspect that have not yet been elicited.
\end{Verbatim}
\end{tcolorbox}

\subsection{Qualitative Examples}
\label{subsect:appendix_examples}

In this section, we present representative examples that qualitatively illustrate the differences between the single-actor baseline and the proposed asymmetric actor–critic framework. These examples highlight how the critic provides real-time supervision that helps the actor avoid errors, enforce constraints, and reach correct decisions during interaction.

All examples compare two systems: (1) a single-actor baseline using \textbf{Claude-4 Sonnet}, and (2) the proposed asymmetric actor--critic framework, \asymacsft\ , which uses the same Claude-4 Sonnet actor supervised by a fine-tuned \textbf{Qwen3-8B} critic.

To demonstrate the generality of the approach, we present examples from multiple domains and benchmarks: one example from $\tau$-bench Retail, one from $\tau$-bench Airline, and three from UserBench (T-22, T-33, and T-44). These examples cover diverse failure modes, including policy violations, incorrect reasoning over user preferences, and suboptimal decision paths, where critic interventions play a crucial role in guiding the actor toward successful outcomes.

\subsubsection{Example ($\tau$-bench, Retail): Critic Guiding Multi-Step Cost Reduction}

\paragraph{Context.}
The user recently placed an order (\#W9348897) containing five items: an action camera (\$481.50), patio umbrella (\$288.82), desk lamp (\$150.01), T-shirt (\$53.27), and backpack (\$193.38), for a total of \textbf{\$1,166.98}.  
However, the user realizes that their credit card has only \textbf{\$1,131} remaining balance and asks whether the payment can be split between two cards.

According to the retail system policy, orders must be paid with a single payment method, so payment splitting is not supported.  
The agent must therefore explore alternative solutions such as modifying the order items or reducing the total cost.

\paragraph{Actor-only baseline.}
In the actor-only baseline, the agent recognizes that payment splitting is not supported. When the user asks to reduce the cost by removing the most expensive item, the actor incorrectly concludes that individual items cannot be removed from a pending order and prematurely suggests canceling the entire order instead. This fails the task because the correct solution is to reduce the order cost without canceling the order.

\paragraph{Actor proposal (before critique).}
\textit{``I understand you'd like to remove the Action Camera to reduce the cost, but unfortunately I don't have the ability to remove individual items from a pending order.  
The options I can help you with are:  
(1) cancel the entire order, or  
(2) change the payment method.''}

\paragraph{Critic feedback (summarized).}
\textit{The current response prematurely leads the conversation toward cancellation.  
Instead, the agent should continue exploring other valid ways to modify the order—such as switching items to cheaper variants—to reduce the total cost while keeping the order active.}

\paragraph{Actor revision.}
Guided by the critic’s feedback, the actor continues exploring the decision space and checks whether cheaper variants of the ordered products exist.  
It identifies lower-cost alternatives for several items:

\begin{itemize}[itemsep=0pt, topsep=0mm, parsep=0pt]
\item Action Camera: \$481.50 $\rightarrow$ \$466.75  
\item Desk Lamp: \$150.01 $\rightarrow$ \$135.24  
\item T-Shirt: \$53.27 $\rightarrow$ \$46.66  
\end{itemize}

The resulting total becomes \textbf{\$1,130.85}, which is below the user’s \$1,131 credit limit.

\paragraph{Outcome.}
After confirming with the user, the actor successfully modifies the order by switching the relevant items to their cheaper variants.  
The order remains active with a final total of \textbf{\$1,130.85}, and the price difference (\$36.13) is refunded to the user’s card. See Table~\ref{tab:example_tau_retail} for a comparison of the outcomes.

\begin{table}[h]
\centering
\caption{Example from $\tau$-bench retail where the critic guides the agent to explore cost-reduction strategies instead of prematurely canceling the order.}
\label{tab:example_tau_retail}
\begin{adjustbox}{width=\linewidth}
\begin{tabular}{lll}
\hline
Method & Final Outcome & Result \\
\hline
Actor Only & Cancels order prematurely & Fail \\
\asymacsft & Modifies items to cheaper variants & Success \\
\hline
\end{tabular}
\end{adjustbox}
\end{table}

\subsubsection{Example ($\tau$-bench, Airline): Critic Enforcing Correct Refund Policy}

\paragraph{Context.}
The user, \textit{James Taylor} (a Silver member), wants to modify an existing round-trip reservation from Las Vegas (LAS) to Houston (IAH). The original outbound itinerary is a one-stop flight (LAS → PHX → IAH). During the conversation, the user requests two modifications:

\begin{itemize}[itemsep=0pt, topsep=0mm, parsep=0pt]
\item Change the outbound itinerary to a \textbf{nonstop flight} from LAS to IAH.
\item \textbf{Remove a checked bag} from the reservation and receive the corresponding refund.
\end{itemize}

After retrieving the reservation details, the agent identifies available nonstop flights and the user selects flight \textbf{HAT266}. During the payment step, the system identifies two stored payment methods in the user’s account:

\begin{itemize}[itemsep=0pt, topsep=0mm, parsep=0pt]
\item Travel certificate: \texttt{certificate\_9380982}
\item Gift card: \texttt{gift\_card\_5634230}
\end{itemize}

According to airline policy, \textbf{travel certificates are non-refundable}, and refunds must be returned to the original payment method.

\paragraph{Actor-only baseline.}
In the baseline system, the actor incorrectly attempts to process the modification using the travel certificate. This violates the airline policy because certificates cannot be used for refunds or modifications in this context, causing the workflow to fail.

\paragraph{Actor proposal (before critique).}
\textit{``Perfect! Now I have all the information needed. Let me proceed with updating your flight reservation using certificate\_9380982.''}

\paragraph{Critic feedback (summarized).}
\textit{The agent is attempting to use a travel certificate for the refund, which violates airline policy. The remaining balance of a certificate is non-refundable, so the refund must be processed using the original payment method instead.}

\paragraph{Actor revision.}
After receiving the critic's feedback, the actor corrects the payment method and processes the modification using the user's gift card instead.

\paragraph{Outcome.}
The system successfully updates the outbound flight to the nonstop option and processes the \$61 refund using \texttt{gift\_card\_5634230}, complying with airline policy.  
In contrast, the actor-only baseline fails due to the incorrect use of the certificate.
See Table~\ref{tab:example_tau_airline} for a comparison of the outcomes.

\begin{table}[h]
\centering
\caption{Example from $\tau$-bench airline where the critic prevents a policy violation in refund processing.}
\label{tab:example_tau_airline}
\begin{tabular}{lll}
\hline
Method & Refund Method & Result \\
\hline
Actor Only & Certificate (invalid) & Fail \\
\asymacsft & Gift card (correct) & Success \\
\hline
\end{tabular}
\end{table}

\subsubsection{Example (UserBench, T-22): Critic Correcting a Suboptimal Rental Car Recommendation}

\paragraph{Context.}
The user plans to rent a car for two drivers and is concerned about losing 
credit-card rental coverage.

\paragraph{Actor-only baseline.}
The actor recommends rental car \textbf{C17}, which does not provide 
belongings waiver insurance. Reward: 0.0.

\paragraph{Actor proposal (before critique).}
\textit{"Based on your travel needs, I recommend rental car C17."}

\paragraph{Critic feedback (summarized).}
\textit{"The recommended option does not include belongings waiver insurance,
even though the user explicitly expressed concern about losing credit-card
coverage. The requirement of two drivers is also not properly considered."}

\paragraph{Actor revision.}
\textit{"I recommend rental car C1 (Hertz Toyota Camry), which supports an
additional driver and includes belongings waiver insurance."}

\paragraph{Outcome.}
The revised recommendation satisfies the user’s requirements and receives
reward 1.0.
See Table~\ref{tab:example_t22} for a comparison of the outcomes.

\begin{table}[h]
\centering
\caption{Example from UserBench T-22 where the critic corrects a suboptimal rental car recommendation.}
\label{tab:example_t22}
\begin{tabular}{ll}
\hline
Method & Rental Car Recommendation \\
\hline
Actor Only & C17 (wrong, reward = 0.0) \\
\asymacsft & C1 (optimal, reward = 1.0) \\
\hline
\end{tabular}
\end{table}

\subsubsection{Example (UserBench, T-33): Critic Correcting an Incorrect Flight Recommendation}

\paragraph{Context.}
The user is planning a trip from New York to Paris.

From the conversation, the user expressed the following preferences:

\begin{itemize}[itemsep=0pt, topsep=0mm, parsep=0pt, ]
\item Travel date: December 15 (one-way flight).
\item Budget conscious: prefers the \textbf{cheapest option that satisfies preferences}.
\item Does \textbf{not like tight layovers} and prefers having more time between flights.
\item Enjoys \textbf{comfortable layovers with time to relax or explore}.
\end{itemize}

\paragraph{Actor-only baseline.}
Without critic supervision, the actor recommends flight \textbf{F4}.  
This option violates the user’s preferences and is not among the optimal choices, resulting in a reward of \textbf{0.0}.

\paragraph{Actor proposal (before critique).}
\textit{``Based on your travel preferences, I recommend flight F4.''}

\paragraph{Critic feedback (summarized).}
\textit{The recommended flight does not satisfy the user’s preferences regarding layover duration and budget considerations.  
Please reconsider the recommendation and select a flight that satisfies the user’s constraints while remaining cost-effective.}

\paragraph{Actor revision.}
Guided by the critic’s feedback, the actor revises its recommendation and selects the correct flight option that satisfies the user’s constraints.

\paragraph{Outcome.}
The revised recommendation satisfies the user’s preferences and is evaluated as the correct choice, resulting in a reward of \textbf{1.0}.  
In contrast, the actor-only system would recommend flight \textbf{F4}, producing an incorrect result with reward \textbf{0.0}. 
See Table~\ref{tab:example_t33} for a comparison of the outcomes.

\begin{table}[h]
\centering
\caption{Example from UserBench T-33 where the critic corrects an incorrect flight recommendation.}
\label{tab:example_t33}
\begin{tabular}{ll}
\hline
Method & Flight Recommendation \\
\hline
Actor Only & F4 (wrong, reward = 0.0) \\
\asymacsft & F16 (optimal, reward = 1.0) \\
\hline
\end{tabular}
\end{table}

\subsubsection{Example (UserBench, T-44): Critic Correcting an Incorrect Apartment Recommendation}

\paragraph{Context.}
The user is planning a trip to Paris with their partner and is searching for an apartment stay. 
During the conversation, the user specifies several preferences for the accommodation:

\begin{itemize}[itemsep=0pt, topsep=0mm, parsep=0pt]
\item The apartment should accommodate \textbf{two guests}.
\item The user prefers an apartment located in a \textbf{central area of the city}.
\item The apartment should have a \textbf{high rating} to ensure a comfortable stay.
\item The user is \textbf{budget-conscious} and prefers the \textbf{lowest-cost option among those satisfying all requirements}.
\end{itemize}

Therefore, the correct choice must satisfy all stated constraints while also being the cheapest feasible option.

\paragraph{Actor-only baseline.}
Without critic supervision, the actor recommends apartment \textbf{A1}.  
However, this option does not satisfy all of the user’s stated preferences and is therefore considered an incorrect recommendation, resulting in a reward of \textbf{0.0}.

\paragraph{Actor proposal (verbatim).}
\textit{“Based on your travel preferences, I recommend apartment A1.”}

\paragraph{Critic feedback (summarized).}
\textit{The recommended apartment does not fully satisfy the user’s preferences regarding location, rating, or cost efficiency. 
Please reconsider the recommendation and select an apartment that satisfies all user constraints and minimizes cost among feasible options.}

\paragraph{Actor revision.}
After receiving the critic’s feedback, the actor reevaluates the available options and selects a different apartment, \textbf{A13}, that satisfies all constraints and is the cheapest among valid choices.

\paragraph{Outcome.}
The revised recommendation A13 satisfies the user’s requirements and is evaluated as the correct option, resulting in a reward of \textbf{1.0}. 
In contrast, the actor-only system would recommend apartment \textbf{A1}, producing an incorrect result with reward \textbf{0.0}. 
% This example illustrates how the critic detects constraint violations and guides the actor toward the correct decision in a multi-turn conversational setting.
See Table~\ref{tab:example_t44} for a comparison of the outcomes.

\begin{table}[h]
\centering
\caption{Example from UserBench T-44 where the critic corrects an incorrect apartment recommendation.}
\label{tab:example_t44}
\begin{tabular}{ll}
\hline
Method & Apartment Recommendation \\
\hline
Actor Only & A1 (wrong, reward = 0.0) \\
\asymacsft & A13 (optimal, reward = 1.0) \\
\hline
\end{tabular}
\end{table}

\section{Data Generation for Critic Training}
\label{sec:appendix_data_gen}

\subsection{Algorithm}
\label{subsec:appendix_data_gen_algorithm}

We present the pesudocode of the critic data generation pipeline in Algorithm~\ref{alg:critic_data_generation}.

\begin{algorithm}[H]
\caption{Critic Data Generation}
\label{alg:critic_data_generation}
\begin{algorithmic}[1]
\REQUIRE Tasks $\{x^{(i)}\}_{i=1}^{N}$ inferred from existing data $\mathcal{D}$, actor $\pi_A$, critic $C_{\theta}$, threshold $\psi$
\ENSURE Critic SFT dataset $\mathcal{S}$
\STATE $\mathcal{S} \leftarrow \emptyset$
\FOR{task $x^{(i)}, i=1,2,\dots,N$}
    \STATE Obtain actor-only trajectories $\{\widehat{\tau}_k\}_{k=1}^{K}$
    \IF{$\sum_{k=1}^K \mathbb{I}[R(\widehat{\tau}_k)< 1] \ge \psi$ (hard task)}
        \STATE Obtain actor-critic trajectories $\{\widetilde{\tau}_k\}_{k=1}^{K}$
            \IF{$R(\widetilde{\tau}_k)=1$, $\forall k\in [K]$ and critic intervenes}
                \STATE { $\mathcal{S} \leftarrow \mathcal{S} \cup 
                \{(x_t, y_t)\mid (x_t, y_t)\in\widetilde{\tau}_k\}$ }
            \ENDIF
    \ENDIF
\ENDFOR
\RETURN $\mathcal{S}$
\end{algorithmic}
\end{algorithm}

\subsection{Additional Statistics}
\label{subsec:appendix_add_statistics}

We report additional statistics of the generated training data used for critic fine-tuning on the two benchmarks. 
The \textit{number of trajectories} refers to the total number of interaction trajectories used to construct the training dataset. 
Since each critic evaluation point (i.e., each turn where the critic evaluates the actor’s proposed action) is treated as one training instance, a single trajectory may produce multiple samples. 
The \textit{number of total samples} therefore denotes the total number of critic training instances extracted from all trajectories. 
Among these, \textit{positive samples} correspond to cases where the critic rejects the actor’s proposed action, while \textit{negative samples} correspond to cases where the critic approves the proposed action.
The resulting dataset statistics for $\tau$-bench and UserBench are summarized in Table~\ref{tab:critic_data_stats}.

\begin{table*}[t]
\centering
\caption{Statistics of the generated critic training data. Each critic evaluation point in a trajectory corresponds to one training sample. Positive samples denote critic rejections, while negative samples denote approvals. For UserBench, the critic is trained on data aggregated across all environments.}
\label{tab:critic_data_stats}
\begin{tabular}{llrrrr}
\hline
Benchmark & Domain/Env & \# Trajectories & \# Samples & \# Positive & \# Negative \\
\hline
$\tau$-bench & Airline & 403 & 2363 & 713 & 1650 \\
$\tau$-bench & Retail & 171 & 1184 & 259 & 925 \\
\hline
UserBench & T-22 & 424 & 848 & 701 & 147 \\
UserBench & T-33 & 345 & 690 & 548 & 142 \\
UserBench & T-44 & 292 & 584 & 460 & 124 \\
UserBench & Total & 1061 & 2122 & 1709 & 413 \\
\hline
\end{tabular}
\end{table*}

\section{More About the Benchmarks}
\label{sec:appendix_benchmarks}

We evaluate the proposed asymmetric actor--critic framework on two multi-turn conversational agent benchmarks: $\tau$-bench~\cite{yao2024tau_bench} and UserBench~\cite{qian2025userbench}. Both benchmarks simulate realistic user-facing environments where agents must interact with users over multiple turns, adhere to domain-specific constraints, and execute structured actions through external tools or databases.

\subsection{$\tau$-bench}

$\tau$-bench~\cite{yao2024tau_bench} evaluates task-oriented conversational agents operating under explicit domain policies. The benchmark simulates customer service scenarios in which the agent interacts with a user and executes actions that modify an underlying system state. Typical tasks include processing refund requests, updating reservations, or canceling flights.

Each task consists of a user query, a structured environment, and a set of domain policies that define valid and invalid actions. The agent must reason over the conversation history, consult the policy rules, and execute appropriate tool calls to complete the task. Actions typically correspond to API calls that update the environment (e.g., booking, canceling, or modifying a reservation).

A key challenge in $\tau$-bench is strict policy compliance. The agent must ensure that its actions satisfy all domain constraints while correctly interpreting user intent. For example, certain refund requests may only be processed under specific conditions, and violating these rules results in task failure.

Performance is evaluated based on task success, which requires the agent to execute the correct sequence of actions while respecting all policy constraints. The benchmark therefore emphasizes reliable decision-making under explicit rule-based constraints.

\subsection{UserBench}

UserBench~\cite{qian2025userbench} evaluates conversational agents in a travel planning environment. In this benchmark, the agent must recommend exactly one option for each travel aspect (e.g., apartment, restaurant, rental car) from a structured database.

Unlike $\tau$-bench, where constraints are explicitly defined by policies, UserBench introduces constraints implicitly through user preferences expressed during the conversation. The agent must interact with the user over multiple turns to elicit both explicit and implicit preferences, such as desired ratings, price limits, or location requirements.

At each step, the agent may ask clarifying questions, search the database, or provide a recommendation. A recommendation is considered optimal if it satisfies all user preferences while minimizing cost among feasible candidates.

This benchmark therefore requires agents to both infer user preferences and ensure that final recommendations satisfy all constraints derived from the dialogue. Success depends on accurately interpreting user intent, gathering missing information, and selecting the optimal option from the available candidates.

\subsection{Evaluation Setting}

Both benchmarks operate in a multi-turn conversational setting in which the agent must interact with a simulated user and environment. 

The benchmarks differ in the nature of their constraints. $\tau$-bench focuses on explicit policy compliance in customer service environments, while UserBench focuses on implicit constraints derived from user preferences in travel planning tasks. Together, they provide complementary testbeds for evaluating the reliability and controllability of conversational agents.

\section{Adapting CRITIC~\cite{gou2024CRITIC} to Multi-Turn Conversational Tasks}
\label{sec:appendix_baselines}

All baselines use the native benchmark implementations or directly report results from the benchmark papers, except CRITIC~\cite{gou2024CRITIC}. 
We adapt and re-implement CRITIC for our one-shot multi-turn conversational setting because the original framework was primarily designed for simpler single-step tasks such as free-form question answering (QA), program synthesis, or toxicity reduction.

\paragraph{Overview of CRITIC.}
CRITIC is a self-correction framework that enables large language models to iteratively verify and revise their own outputs using external feedback. 
The method first generates an initial answer using the base model, then invokes external tools (e.g., search engines, code interpreters, or evaluation APIs) to critique certain aspects of the generated output. 
Based on the feedback returned by these tools, the model revises its response to produce an improved final output. 
This generate--critique--revise loop allows the model to correct factual errors, flawed reasoning, or undesirable outputs by leveraging external verification signals rather than relying solely on its internal knowledge. 

\paragraph{Challenges in Multi-Turn Conversational Settings.}
Directly applying CRITIC to our setting presents several challenges. 
First, the tasks in $\tau$-bench and UserBench involve long multi-turn interactions where the agent must maintain dialogue context, reason about evolving user goals, and adhere to domain constraints across multiple steps. 
In contrast, CRITIC was originally evaluated on tasks where the model generates a single output for a given prompt, making the critique process relatively localized.

Second, error patterns in multi-turn conversations are substantially more complex than those in, for example, typical QA settings. 
In QA, common failures such as factual hallucinations can often be detected by verifying a single generated statement. 
However, conversational agents may fail due to a variety of factors, including incorrect policy reasoning, missing user preference information, incorrect tool usage, or compounding errors from earlier turns in the dialogue. 
These failure modes are harder to capture with simple critique templates.

Finally, CRITIC relies heavily on demonstration examples that illustrate how the model should critique and revise its outputs. 
Constructing such demonstrations is significantly more challenging in long conversational tasks because the relevant error may originate from earlier turns, or from misunderstandings of implicit user preferences.

\paragraph{Construction of Demonstration Examples.}
To adapt CRITIC to our setting, we construct demonstration examples that focus on localized decision points within a dialogue rather than the entire conversation trajectory. 
Specifically, we identify turns where the actor proposes an action that either leads to an incorrect outcome or corresponds to a desired correct behavior. 
Each demonstration consists of three components: (1) the dialogue context leading up to the actor’s proposal, (2) the actor’s proposed action, and (3) a critique explaining whether the proposal is correct and how it should be revised if necessary.

These demonstrations are designed to capture common conversational patterns that occur in our benchmarks. 
For $\tau$-bench, we include examples illustrating policy violations (where the actor approves an illegal action), policy hallucinations (where the actor denies a valid request due to non-existent constraints), and desired actions that correctly follow domain policies. 
For UserBench, we include cases where the actor selects options that violate expressed user preferences, chooses non-optimal options that are not the cheapest among feasible candidates, or fails to reason over incomplete preference information, as well as examples demonstrating correct optimal recommendations.

By focusing critiques on specific decision points rather than entire trajectories, we make it feasible to apply CRITIC-style self-correction in long multi-turn interactions. 
To ensure a fair comparison, we align the self-correction mechanism with the intervention design of our asymmetric actor--critic framework in two ways. 
First, self-correction is invoked at exactly the same dialogue turns where the critic would intervene in the proposed asymmetric actor--critic system. 
Second, since the asymmetric actor--critic revises the actor's proposal at most once per intervention point, we also allow CRITIC to perform only a single self-correction step at each such point. 
Nevertheless, designing effective demonstration examples for conversational settings remains challenging, as errors may originate from earlier dialogue turns, evolving user preferences, or complex policy reasoning. 
This difficulty highlights an inherent limitation of CRITIC when applied to long multi-turn interactions.

\section{Broader Impact and Potential Risks}
\label{sec:appendix_risks_broader_impact}

Improving the reliability and controllability of LLM agents may benefit real-world applications such as customer support, travel planning, and decision assistance systems. By enabling lightweight critics to supervise stronger proprietary actors, our approach provides a practical way to improve agent reliability without modifying the underlying model. This may help organizations deploy LLM-based agents more reliable in user-facing environments.

However, automated conversational systems must still be deployed with appropriate safeguards and human oversight. Incorrect critic interventions or failures to detect errors could still lead to undesirable outcomes, and careful monitoring is necessary when applying such systems in high-stakes environments.

\section{Use of AI Assistants}

AI-based writing assistants were used for language editing, drafting support, and refinement of the teaser figure.
All technical content, experimental design, and analysis were produced and verified by the authors.

\end{document}